\documentclass[lettersize,journal]{IEEEtran}
\usepackage{amsmath,amsfonts}
\usepackage{algorithmic}
\usepackage{algorithm}
\usepackage{array}
\usepackage[caption=false,font=normalsize,labelfont=sf,textfont=sf]{subfig}
\usepackage{textcomp}
\usepackage{stfloats}
\usepackage{url}
\usepackage{verbatim}
\usepackage{graphicx}
\usepackage{multirow}
\usepackage{cite}
\usepackage{color}
\usepackage{hyperref}
\usepackage{amssymb}
\usepackage{colortbl}

\DeclareRobustCommand\onedot{\futurelet\@let@token\@onedot}

\def\etc{\emph{etc}}  
            
\def\etal{\emph{et al.} }
\makeatother

\begin{document}
    
\title{Towards Real-World Blind Face Restoration with Generative Diffusion Prior}

\author{Xiaoxu Chen,
        Jingfan Tan,
        Tao Wang,
        Kaihao Zhang,
        Wenhan Luo,
        Xiaochun Cao\\
\IEEEcompsocitemizethanks{\IEEEcompsocthanksitem X. Chen, J. Tan, W. Luo and X. Cao are with Shenzhen Campus of Sun Yat-sen University, Shenzhen, China, 
(e-mail: \{chenxiaoxu89, tjfky2001, whluo.china\}@gmail.com, caoxiaochun@mail.sysu.edu.cn).
\IEEEcompsocthanksitem T. Wang is with the State Key Laboratory for Novel Software Technology, Nanjing University, Nanjing, China, (e-mail: taowangzj@gmail.com).
\IEEEcompsocthanksitem K. Zhang is with the College of Engineering and Computer Science, Australian National University, Canberra, Australia, (e-mail: \{super.khzhang\}@gmail.com).

}

}


\maketitle
\begin{abstract}
Blind face restoration is an important task in computer vision and has gained significant attention due to its wide-range applications. Previous works mainly exploit facial priors to restore face images and have demonstrated high-quality results. However, generating faithful facial details remains a challenging problem due to the limited prior knowledge obtained from finite data. In this work, we delve into the potential of leveraging the pretrained Stable Diffusion for blind face restoration. We propose BFRffusion which is thoughtfully designed to effectively extract features from low-quality face images and could restore realistic and faithful facial details with the generative prior of the pretrained Stable Diffusion. In addition, we build a privacy-preserving face dataset called PFHQ with balanced attributes like race, gender, and age. This dataset can serve as a viable alternative for training blind face restoration networks, effectively addressing privacy and bias concerns usually associated with the real face datasets. Through an extensive series of experiments, we demonstrate that our BFRffusion achieves state-of-the-art performance on both synthetic and real-world public testing datasets for blind face restoration and our PFHQ dataset is an available resource for training blind face restoration networks. The codes, pretrained models, and dataset are released at \href{https://github.com/chenxx89/BFRffusion}{https://github.com/chenxx89/BFRffusion}.
\end{abstract}
 

\section{Introduction}
In real-world scenarios, face images may suffer from various types of degradation, such as noise, blur, down-sampling, JPEG compression artifacts, and \etc. Blind face restoration aims to restore high-quality face images from low-quality ones that suffer from unknown degradation. Due to its extensive range of applications, blind face restoration has gained significant attention in the field of computer vision.

Because of the unique structural and semantic information of face images, previous blind face restoration methods typically exploit face priors to restore face images, such as reference prior \cite{DFDNet, dmdnet, Restoreformer}, geometric prior \cite{Fsrnet, kim2019progressive, yu2018face, PSFRGAN}, and generative prior \cite{Pulse, GFPGAN, GPEN, DEARGAN}. Specifically, reference prior-based methods usually employ the facial structure \cite{DFDNet,dmdnet} or facial component dictionary \cite{Restoreformer} obtained from additional high-quality face images as the reference prior to guide the face restoration process. The unique geometric shapes and spatial distribution information of faces in the images are utilized to gradually recover high-quality face images in geometric prior-based methods. Geometric prior mainly include facial landmarks \cite{Fsrnet,kim2019progressive}, facial heatmaps \cite{yu2018face}, and facial parsing maps \cite{PSFRGAN}. With the development of generative adversarial networks (GANs), researchers have started to leverage the generative prior for face image restoration. Generative prior typically includes GAN inversion \cite{Pulse} and pretrained facial GAN models \cite{GFPGAN, GPEN, DEARGAN} to provide richer and more diverse facial information. However, the prior knowledge derived solely from limited data may be not enough for restoration purposes.

\begin{figure}[t] 
\centering
\includegraphics[width=0.48\textwidth]{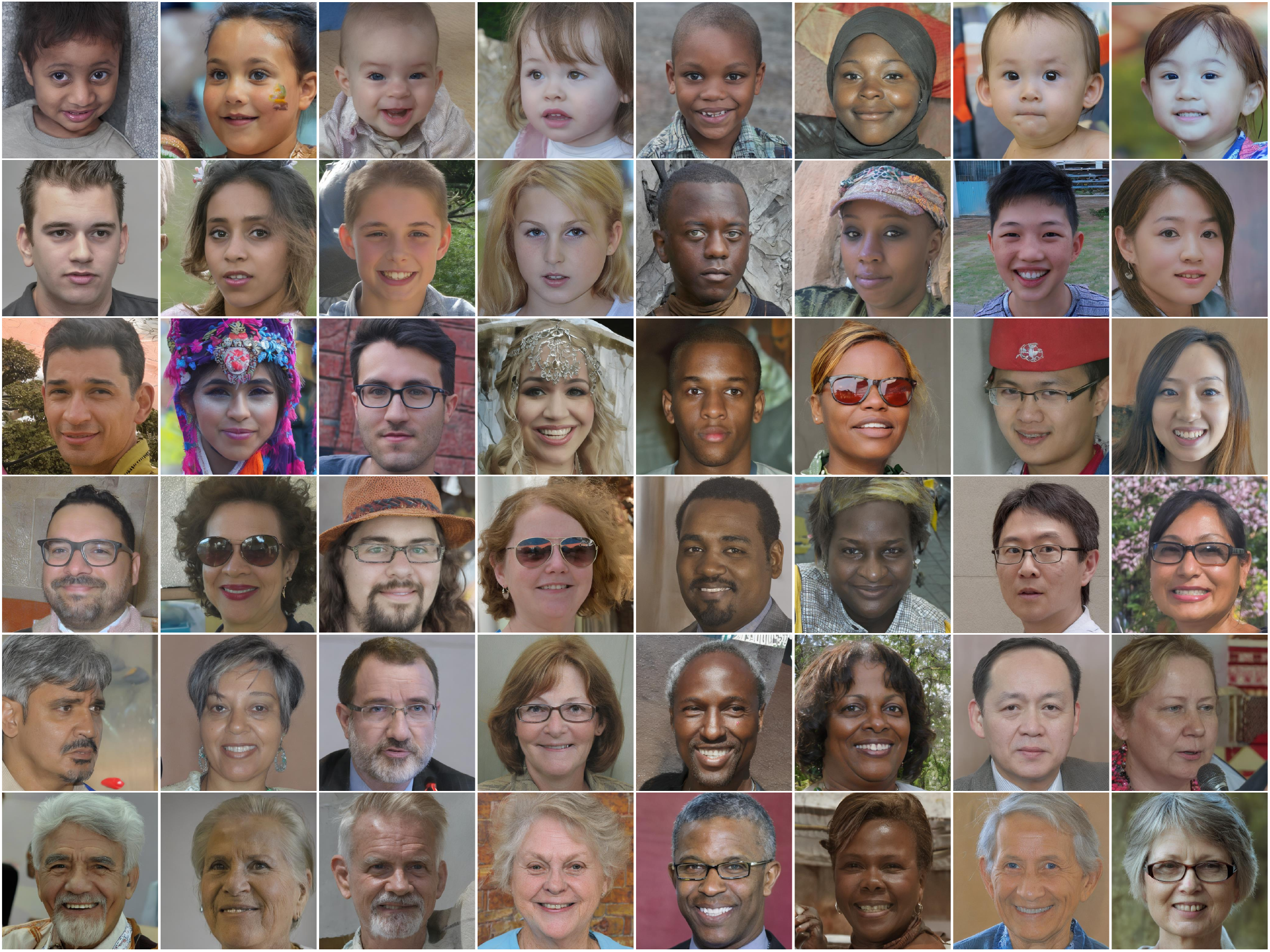}
\caption{Representative face images from the proposed PFHQ dataset. These face images exhibit balanced race, gender, and age distribution.}
\label{dataset}
\end{figure}

Recently, the diffusion model has achieved remarkable results in image-generation tasks. The diffusion model is a two-stage generative model comprising a forward diffusion stage and a reverse denoising stage. During the forward diffusion stage, the input is gradually added with Gaussian noise and eventually is transformed into a random noise that conforms to the Gaussian distribution. In the reverse denoising stage, the model reconstructs the original input data distribution by denoising step by step. The diffusion model is highly regarded for its ability to produce samples with exceptional quality and has also received extensive attention in all kinds of image restoration tasks. SR3 \cite{SR3} adapts the diffusion model to image super-resolution through a stochastic denoising process. Xia \etal \cite{xia2023diffir} use the diffusion model to estimate a compact image restoration prior to guide restoration, and achieve state-of-the-art performance with lower computational costs. In order to reduce computational requirements, the latent diffusion model \cite{latentdiffusion} applies the diffusion and denoising process in the latent space, and Stable Diffusion is a large-scale implementation of it. Stable Diffusion is trained on billions of text-image pairs and exhibits a powerful image-generation ability. StableSR \cite{stalesr} and DiffBIR \cite{diffbir}  achieve realistic image restoration performance with the help of the generative ability of Stable Diffusion.

In this paper, we further explore the generative ability of the pretrained Stable Diffusion in the field of blind face restoration. Compared with the GAN prior used in previous generative prior-based methods \cite{GFPGAN, GPEN, DEARGAN}, the pretrained Stable Diffusion can provide richer and more diverse priors including facial components and general object information, making it possible to generate realistic and faithful facial details. However, Stable Diffusion is a text-to-image generation model and can not be applied to restoration tasks directly without any modification. Although StableSR \cite{stalesr} and DiffBIR \cite{diffbir} leveraging the priors use existing U-net architecture of Stable Diffusion to extract image features, such design limits the restoration performance and efficiency. To address this, we propose BFRffusion with delicately designed architecture to leverage generative priors encapsulated in the pretrained Stable Diffusion for blind face restoration. Specifically, our BFRffusion consists of four modules: shallow degradation removal module (SDRM), multi-scale feature extraction module (MFEM), trainable time-aware prompt module (TTPM), and pretrained denoising U-Net module (PDUM). SDRM removes the shallow degradation of the input and encodes it into latent space. MFEM adopts transformer blocks to extract multi-scale features added to the pretrained U-Net to guide the restoration process. TTPM generates the time-aware prompt which can provide semantic guidance for the restoration in different time steps. PDUM serves as the core denoising network of the diffusion model. Compared with previous methods \cite{stalesr, diffbir} based on Stable Diffusion, more effective transformer blocks and time-aware prompts are applied in our BFRffusion. What is more, we employ a distinctive training strategy for the PDUM. Comprehensive experiments show that our BFRffusion achieves state-of-the-art performance on both synthetic and real-world datasets.

Previous works \cite{DFDNet, PSFRGAN, GFPGAN, VQFR, Restoreformer, codeformer} mainly train the restoration networks on Flickr-Faces-HQ (FFHQ) \cite{ffhq} dataset. This dataset consists of 70K high-quality aligned face images collected from the Internet. However, real face datasets suffer from privacy issues and racial bias problems. What is more, collecting a high-quality dataset is a costly process. Recently, synthesis face datasets have received increasing attention in computer vision, and researchers \cite{SyntheticData1, SyntheticData2, SyntheticData3, SyntheticData4, zhao2017dual} have started utilizing these datasets for training neural networks. To address privacy and bias concerns of the real face datasets, we provide a synthetic face dataset called Privacy-preserving-Faces-HQ (PFHQ) with balanced race, gender, and age for training the restoration networks. Representative face images of our proposed PFHQ dataset are shown in Fig. \ref{dataset}. To obtain the dataset, we first employ ControlNet \cite{ControlNet} which could add conditional information to the pretrained Stable Diffusion to generate a large number of face images. Then we select and classify the synthetic face images. Finally, we adopt a widely used degradation model to synthesize low-quality face images. Meanwhile, we build PFHQ-Test using the same operations to evaluate the performance of restoration networks on balanced data. Extensive experiments show that our PFHQ dataset achieves comparable performance in blind face restoration tasks compared to the widely used FFHQ dataset. We expect our PFHQ dataset can drive the development of face restoration and other face-related tasks in the field of computer vision.

Overall, we summarize the contributions as follows:
\begin{itemize}
    \item We leverage the generative prior encapsulated in the pretrained Stable Diffusion for blind face restoration. The prior which contains abundant facial components and general object information is one of the prerequisites enabling us to restore realistic and faithful facial details.
    \item We propose a blind face restoration method called BFRffusion with delicately designed architecture that could effectively extract multi-scale features from low-quality face images and sufficiently leverage the diffusion prior.
    \item According to extensive experimental studies, our BFRffusion achieves state-of-the-art performance compared with existing methods on both synthetic and real-world datasets.
    \item We provide a privacy-preserving and balanced dataset called PFHQ that includes 60K paired face images for training blind face restoration networks. This dataset achieves utility on par with the widely used FFHQ dataset.
\end{itemize}

The following sections of this paper are organized as follows: Section \ref{Related Work} presents a review of relevant research covering diffusion models in image restoration, blind face restoration, and face datasets. Section \ref{Methodology} details the architecture of our BFRffusion model and the construction process of our PFHQ dataset. Section \ref{Experiment} reports the experimental results and their analyses. Finally, Section \ref{Conclusion} presents our conclusions and outlines future work.

\section{Related Work} \label{Related Work}
In this section, we provide a concise overview of diffusion models in image restoration, blind face restoration, and face datasets.

\subsection{Diffusion models in Image Restoration}
The diffusion model demonstrates superior capabilities in generating a more accurate target distribution than other generative models and has achieved excellent results in sample quality. Recently, due to the more stable generation ability than GAN \cite{gan}, the diffusion model has also received extensive attention in all kinds of low-level image restoration tasks. Based on generation space, diffusion-based image restoration methods can be classified into two groups: image space and latent space. Image-based methods generate the structures and textures directly. Latent-based methods utilize a well-designed encoder to transform the images into a compact latent space for generation, thereby improving the generation efficiency. 

The majority of diffusion-based methods generate the restored images within image space. In the image super-resolution task, SR3 \cite{SR3} adapts the diffusion model to generate conditional images and performs super-resolution through a stochastic denoising process. In the image deblurring field, Whang \etal \cite{whang2022deblurring} present an alternative framework for blind deblurring based on conditional diffusion models. Ren \etal \cite{ren2023multiscale} introduce effective multi-scale structure guidance as an auxiliary prior to the image-conditional diffusion model for a significant improvement of the deblurring results. For the image inpainting task, Repaint \cite{lugmayr2022repaint} employs a pretrained unconditional DDPM \cite{DDPM} as the generative prior and alters the reverse diffusion iterations by sampling the unmasked regions using the given image information. In addition, the diffusion model is also used in other image restoration tasks like denoising \cite{feng2023score}, lowlight enhancement \cite{yi2023diff,wang2023exposurediffusion}, shadow removal \cite{guo2023shadowdiffusion}, JPEG artifact removal \cite{saharia2022palette}, and so on.

The latent diffusion model \cite{latentdiffusion} proposes to implement diffusion-based generation in latent space to alleviate the training and sampling costs of the diffusion model. DiffIR \cite{xia2023diffir} exploits the latent-wise diffusion model to generate the compact image restoration priors, which guides the restoration network to achieve better performance. StableSR \cite{stalesr} and DiffBIR \cite{diffbir} achieve realistic image restoration leveraging the generative ability of the pretrained Stable Diffusion which is based on the latent diffusion model. Our BFRffusion also leverages Stable Diffusion, but we apply delicate design and a distinctive training strategy.

\subsection{Blind Face Restoration}
Face restoration is a specific type of image restoration and the restored faces can be used for face recognition \cite{tu2021joint, wang2021face, zhao2018towards}, face detection \cite{chen2021yolo, feihong2023toward} and \etc. Blind face restoration (BFR) aims to restore low-quality face images without any knowledge of degradation \cite{WangTaosurvey}. Compared with general images, the face image typically carries more structural and semantic information. Thus, most blind face restoration methods use facial priors to restore face images with clearer facial details.

Early attempts utilize reference prior \cite{DFDNet,dmdnet, asff, inpaint2} and geometric prior \cite{Fsrnet, kim2019progressive, FSR2, bulat2018super, yu2018face, PSFRGAN} to guide the face restoration process. The geometric prior-based method usually extracted priors like facial landmarks \cite{Fsrnet, kim2019progressive}, facial heatmaps \cite{yu2018face}, and facial parsing maps \cite{PSFRGAN} from low-quality face images. This also causes the restoration effect to be related to the degree of degradation. Reference prior-based methods \cite{DFDNet, asff, dmdnet,tan2023blind} guide the face restoration process by additional high-quality face images. Although this reduces the limitations of geometric priority, it may change the identity information of the restored face. 

With the rapid development of generative networks, pretrained GAN-based models have become the most popular approach in the field of blind face restoration. Many BFR approaches \cite{Pulse, DEARGAN, GPEN, SGPN} use the information encapsulated in the well-trained high-quality face generator as the generative prior to guiding the face restoration process. Several works \cite{GFPGAN, GPEN, SGPN} first extract facial information from the input low-quality face images and find the closest latent vector in the StyleGAN \cite{stylegan2} span as facial prior leveraging the pretrained StyleGAN as the decoder. The works \cite{Restoreformer, VQFR, codeformer} adopt VQGAN \cite{vqgan} to pretrain a high-quality discrete feature codebook on HQ face images as prior. Compared to the StyleGAN prior, the discrete codebook prior, acquired within a smaller agent space, significantly reduces uncertainty. 

Recently, the diffusion model has been proven to be more stable than GAN \cite{gan}, and the generating images are more diverse. This has also received attention in the blind face restoration task. DR2 \cite{wang2023dr2} diffuses input images to a noisy status where various types of degradation give way to Gaussian noise, and then captures semantic information through iterative denoising steps. DiffBIR \cite{diffbir} and our BFRffusion both leverage the pretrained Stable Diffusion as the generative prior which can provide more prior knowledge than other existing methods. Different from DiffBIR, we use delicately designed transformer architecture to extract multi-scale features and design a time-aware prompt module to replace the original CLIP used in DiffBIR to reduce the computational complexity.
 
\subsection{Face Datasets}

\begin{table}[t]
    \centering
    \tabcolsep=0.1cm
    \caption{Representative face datasets. Most of the current public face datasets do not consider the bias problem and they are real face datasets which may cause privacy issues.}
    \begin{tabular}{c|cccc|ccc}
    \hline
        \multirow{2}*{Dataset} & \multirow{2}*{Size}  & \multirow{2}*{Real} & \multirow{2}*{Paired} & \multirow{2}*{HR} & \multicolumn{3}{c}{Balanced}  \\ 
        ~ & ~ & ~ & ~ & ~ & Race & Age & Gender \\ \hline
        LFW \cite{LFW} & 13K & \(\checkmark\) & \(\times\) & \(\times\) & \(\times\)& \(\times\)& \(\times\)\\ 
        CelebA \cite{celeba} & 200K & \(\checkmark\) & \(\times\) & \(\times\) & \(\times\)& \(\times\)& \(\times\)\\ 
        CelebA-HQ \cite{progressivegan} & 30K & \(\checkmark\) & \(\times\) & \(\checkmark\) & \(\times\)& \(\times\)& \(\times\)\\ 
        FFHQ \cite{ffhq} & 70K & \(\checkmark\) & \(\times\) & \(\checkmark\) & \(\times\)& \(\times\)& \(\times\)\\ 
        EDFace-Celeb-1M \cite{edface} & 1.7M & \(\checkmark\) &\(\checkmark\) & \(\times\) & \(\checkmark\) & \(\checkmark\)& \(\times\)\\ 
        FairFace \cite{fairface} & 108K & \(\checkmark\) & \(\times\) & \(\times\) & \(\checkmark\) & \(\checkmark\)& \(\checkmark\)\\ 
        SFHQ \cite{SFHQ} & 425K & \(\times\)  & \(\times\) & \(\checkmark\) & \(\times\)& \(\times\)& \(\times\)\\ \hline
        \textbf{PFHQ (Ours)} & 60K & \(\times\)  & \(\checkmark\) & \(\checkmark\) & \(\checkmark\)& \(\checkmark\)& \(\checkmark\)\\ \hline
    \end{tabular}
    \label{dataset_table}
\end{table}

In this section, we provide an overview of the widely used face datasets constructed recently. The Labeled Faces in the Wild (LFW) dataset \cite{LFW} designed for unconstrained face recognition was collected from the web in 2007. It contains 13,233 face images labeled with the name of the person pictured. The CelebFaces Attributes (CelebA) dataset \cite{celeba} released in 2014 is a large-scale face attributes dataset with 202,599 face images of 10,177 celebrities and is widely used for image generation, image super-resolution, \etc. In order to meet the needs of high-resolution image generation, Karras \etal \cite{progressivegan} created CelebA-HQ, which is a high-quality version of the CelebA dataset \cite{celeba} and consists of 30K face images at \(1024^{2}\) resolution. The Flickr-Faces-HQ (FFHQ) \cite{ffhq} dataset consists of 70K high-quality face images at \(1024^{2}\) resolution and exhibits notable diversity in age, ethnicity, and image backgrounds. 

In recent years, researchers in the field of computer vision have been increasingly focused on the race bias problem. However, a significant challenge lies in most existing datasets, which often exhibit pronounced biases towards specific races. Consequently, blind face restoration models trained on such data may inadvertently produce restored face images that convey inappropriate race-related information. The FairFace \cite{fairface} dataset addresses this issue by offering a balanced collection of 108,501 face images across different races, genders, and age groups. The current blind face restoration methods mainly employ supervised learning and paired training face images are necessarily required. EDFace-Celeb-1M \cite{edface} is a public ethnically diverse face dataset, comprising 1.5M paired face images at \(128^{2}\) resolution and 200K real-world low-resolution images for qualitative testing. However, it is worth noting that the resolutions of FairFace and EDFace-Celeb-1M datasets are much lower than the widely used \(512^{2}\) resolution in current blind face restoration methods \cite{DFDNet, PSFRGAN, GFPGAN, VQFR, Restoreformer, codeformer}.

\begin{figure*}[t]
\centering
\includegraphics[width=0.95\textwidth]{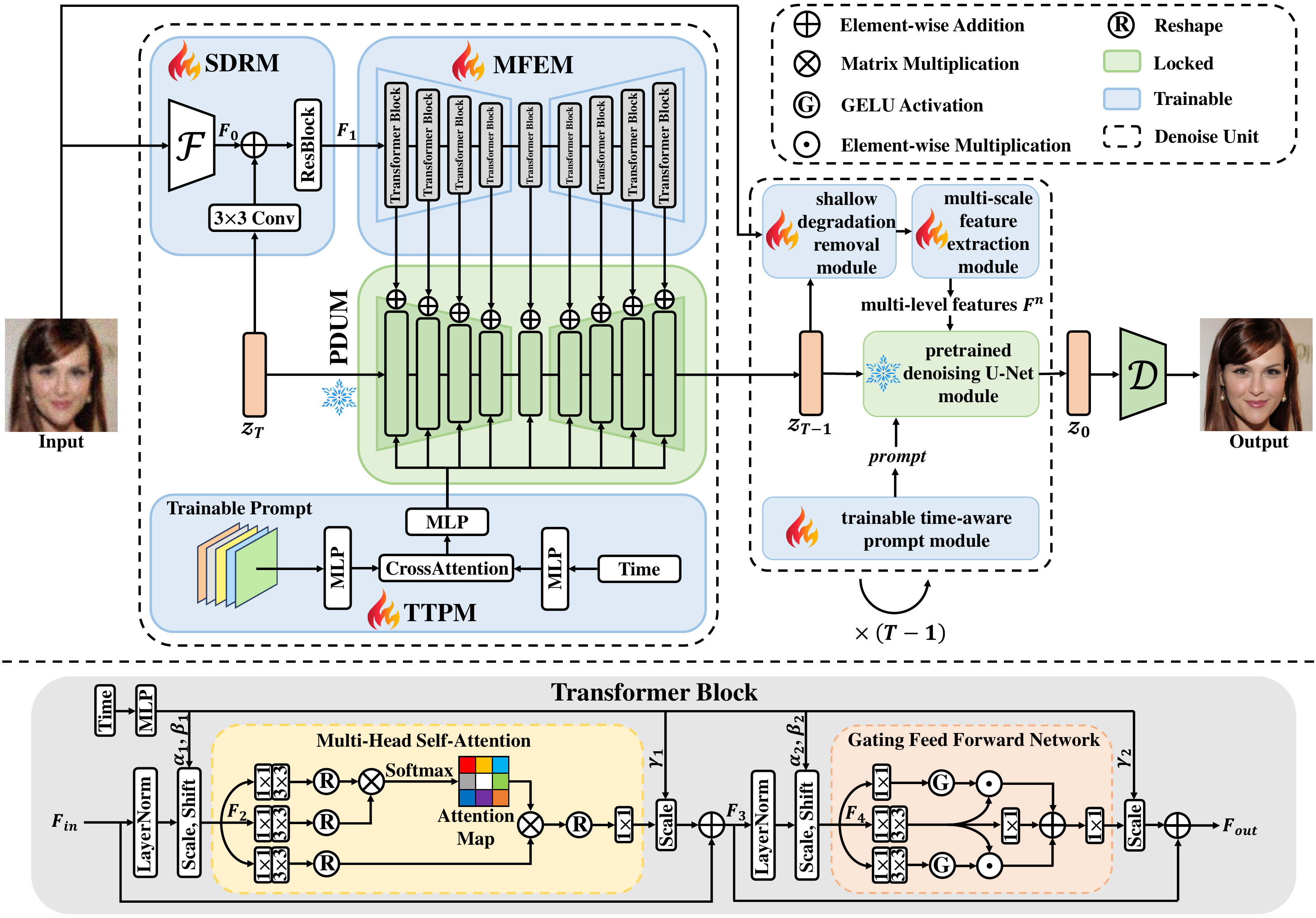}
\caption{Overview of the architecture of BFRffusion which consists of four modules. The shallow degradation removal module (SDRM) and the multi-scale feature extraction module (MFEM) remove shallow degradation and extract multi-scale features from low-quality face images. The pretrained denoising U-Net module (PDUM) utilizes multi-scale features and prompts from the trainable time-aware prompt module (TTPM) as conditions to predict the next step of noise based on the input noise. After multiple denoising steps, high-quality latent features are obtained, which are subsequently transformed into high-quality face images by the pretrained decoder. The MFEM is composed of several transformer blocks, whose structure is illustrated below the dashed line.}
\label{model}
\end{figure*}

With the development of Artificial Intelligence Generated Content (AIGC), particularly the remarkable performance of diffusion models in the field of image generation in recent years, researchers \cite{SyntheticData1, SyntheticData2, SyntheticData3, SyntheticData4, zhao2017dual} have started utilizing synthetic face datasets to train neural networks. The Synthetic Faces High Quality (SFHQ) Dataset \cite{SFHQ}, consisting of approximately 425K synthetic images, was created by encoding the inspiration images into StyleGAN2 \cite{stylegan2} latent space and then manipulating each image to make them appear photo-realistic. However, SFHQ is not paired and balanced. In order to address the ethical and imbalanced issues present in real face datasets and provide convenience to future blind face restoration research, we provide a privacy-preserving and paired face dataset with balanced race, gender, and age. Table \ref{dataset_table} provides a summary of current face datasets to provide a comprehensive overview.

\section{Methodology} \label{Methodology}

\subsection{Blind Face Restoration Method with Generative Diffusion Prior}

\subsubsection{Overall Architecture}
As shown in Fig. \ref{model}, the proposed BFRffusion comprises four modules: shallow degradation removal module (SDRM), multi-scale feature extraction module (MFEM), trainable time-aware prompt module (TTPM, and pretrained denoising U-Net module (PDUM). Specifically, given a degraded face image  \(x \in \mathbb{R} ^{H\times W\times 3} \), the SDRM consisting of several convolutions, activation functions, and a ResBlock \cite{resblock, residual}, first encodes the input image \(x\) into a latent representation \(z \in \mathbb{R} ^{\frac{H}{8}\times \frac{W}{8} \times C} \) and extracts features \(F_{1}\) from it. Then, the features \(F_{1}\) are processed by the MFEM to extract multi-scale features that are appropriate for the different resolutions of Stable Diffusion. The MFEM is composed of several specially designed transformer blocks \cite{attention}. The TTPM constructed with a trainable parameter, a cross-attention block, and several multi-layer perceptron layers (MLP), generates the \(Prompt\) that can guide the restoration process in different time steps. Finally, we add the output features \(F^{n}\) from the MFEM to the PDUM to guide the denoising process and map the \(Prompt\) from the TTPM via a cross-attention layer to provide semantic guidance. A clear latent image is obtained from random Gaussian noise by gradual denoising and can be decoded to a clear image with the decoder of the pretrained VAE \cite{vae}. We provide a detailed explanation of the four modules in our BFRffusion in the subsequent sections.

\subsubsection{Shallow Degradation Removal Module}
The low-quality images of blind face restoration usually suffer from multifarious and complicated types of degradation (e.g., blur, noise, JPEG compression artifacts, low-resolution, \etc). So we propose the shallow degradation removal module (SDRM) to obtain clear latent features from the input low-quality images. Stable Diffusion applies the denoising process in the latent space to reduce the computational resources. Specifically, Stable Diffusion utilizes a pretrained variational autoencoder (VAE) \cite{vae} model with KL loss to encode \(512\times512\) pixel images into \(64\times64\) latent images. In order to match the denoising resolution of Stable Diffusion, we design a simple encoder network \(\mathcal{F} \left ( \cdot  \right ) \) that contains several convolution layers with \(3\times3\) kernels, \(2\times2\) strides, and the Sigmoid Linear Unit (SiLU) function. The network could remove shallow degradation of the input low-quality images and encode them into \(64\times64\) latent images. The encoder network can be expressed as follows:
\begin{equation}
F_{0} = \mathcal{F} \left ( x \right ), 
\label{eq1}
\end{equation}
where \(x\) represents low-quality images and \(F_{0}\) represents \(64\times64\) latent images. Since the Stable Diffusion operates in the noise prediction mode, meaning that the whole diffusion model works on noise, the output of the denoising U-Net is the added noise rather than clear denoised images. So we add the latent images \(F_{0}\) to the randomly sampled noise \(z_{t} \) (\(t\in [1, T]\), where T is the number of diffusion steps) to stabilize the denoising process. We utilize a convolution layer with \(3\times3\) kernels to adjust the strength of the noise \(z_{t} \). What's more, by adding noise of different diffusion steps to latent images, shallow degradation can be mitigated and it can prepare for extracting facial features in the next stage. The inclusion of the time condition is crucial in diffusion models, so we first encode the time step \(t\) using several MLP layers as the time embedding and then add it to the ResBlock.The formulation is as follows:
\begin{equation}
emb = MLP(t),
\label{eq2}
\end{equation}
\begin{equation}
F_{1} = ResBlock((F_{0} + Conv(z_{t})), emb),
\label{eq3}
\end{equation}
where \(F_{1}\) is the output features of SDRM.

\subsubsection{Multi-scale Feature Extraction Module} 
Stable Diffusion is a typical U-Net architectural model that operates at four different resolutions: \(64\times64\), \(32\times32\), \(16\times16\), and \(8\times8\). To extract clear latent features from \(F_{1}\) and align with the resolutions of Stable Diffusion, we propose a transformer-based U-Net called multi-scale feature extraction module (MFEM). Our MFEM comprises several novel transformer blocks that consider both latent feature information and time conditions. We apply adaptive normalization operations \cite{AdaIN, spade, scalablediffusion} in the proposed transformer block to embed the time conditions. Specifically, we generate six affine transformation parameters \(\alpha _{1},\beta _{1},\gamma _{1},\alpha _{2},\beta _{2},\gamma _{2} \) from the time embedding \(emb\) in Eq. \eqref{eq2} using several MLP layers. After that, we apply Spatial Feature Transform (SFT) \cite{SFT} to modulate the input image features that have undergone LayerNorm layer processing. It is formulated as follows:
\begin{equation}
\alpha _{1},\beta _{1},\gamma _{1},\alpha _{2},\beta _{2},\gamma _{2}  = MLP(emb),
\label{eq4}
\end{equation}
\begin{equation}
\begin{aligned}
F_{2} &= SFT(LayerNorm(F_{in})\mid \alpha _{1},\beta _{1}) \\
        &= \alpha _{1} \odot (1 + LayerNorm(F_{in})) + \beta _{1},
\end{aligned}
\label{eq5}
\end{equation}
where \(F_{in}\) is the input features of the proposed transformer block, and \(F_{2}\) represents the output features of the first SFT \cite{SFT}. Then a Multi-Head Self-Attention is employed to capture both global and local contextual information from the input latent features. Initially, we utilize pixel-wise convolutions \(W_{p}\) \cite{pixelconv} and depth-wise convolutions \(W_{d}\) to generate query, key, and value from the output features \(F_{2}\). Then, we perform self-attention \cite{attention} across channels to produce an attention map that encodes the global context implicitly. Following the generation of the attention map, we scale the output of Multi-Head Self-Attention using the affine transformation parameters (\(\gamma _{1}\)). The Multi-Head Self-Attention and scaling process can be formulated by:
\begin{equation}
K, Q, V= W_{p}W_{d}(F_{2}),
\label{eq6}
\end{equation}
\begin{equation}
F_{3} = F_{in} + \gamma _{1} \odot (W_{p}(softmax(KQ/\alpha)V),
\label{eq7}
\end{equation}
where \(F_{3}\) is the scaled output features of Multi-Head Self-Attention and \(\alpha\) is a learnable scaling factor to adjust the dot product of \(K\) and \(Q\). In the subsequent step, we apply the SFT \cite{SFT} method again (similar to Eq. \eqref{eq5}) to scale and shift the features \(F_{3}\) using the affine transformation parameters (\(\alpha _{2},\beta _{2}\)). This operation can be formulated as:
\begin{equation}
\begin{aligned}
F_{4} &= SFT(LayerNorm(F_{3})\mid \alpha _{2},\beta _{2}) \\
        &= \alpha _{2} \odot (1 + LayerNorm(F_{3})) + \beta _{2},
\end{aligned}
\label{eq8}
\end{equation}

Finally, we introduce the Gating Feed Forward Network (GFFN) as a gating mechanism in the feed-forward network to enhance the expressive capacity of networks. Particularly, the GFFN is structured as the element-wise product of three parallel paths, which are composed of pixel-wise convolution \(W_{p}\), depth-wise convolution \(W_{d}\), and the Gelu function \(G\). Following the GFFN, we proceed with modulation by scaling the output using the affine transformation parameters (\(\gamma _{2}\)). The GFFN and scaling process can be formulated by:
\begin{equation}
F_{out} = F_{3} + \gamma _{2} \odot GFFN(F_{4}),
\label{eq9}
\end{equation}
where \(F_{out}\) is the output features of the proposed transformer block. We utilize the transformer block as the fundamental unit and construct the MFEM by incorporating downsampling, upsampling, and skip connections. To match the different resolutions of Stable Diffusion blocks, we collect the output features of all transformer blocks in our MFEM and apply pixel-wise convolutions \(W_{p}\) initialized with Gaussian weights to adjust the strength of the output features between different transformer blocks. The complete MFEM can be calculated as follows,
\begin{equation}
F^{n} = W_{p}(Transformer^{n}(F_{in}^{n})),
\label{eq10}
\end{equation}
where \(Transformer^{n}(\cdot)\) is the \(n^{th}\) proposed transformer block in our MFEM and \(F_{in}^{n}\) is the input features of our MFEM. \(F^{n}\) represents the output features of our MFEM which match the different resolutions of Stable Diffusion blocks.

\begin{figure}[t] 
\centering
\includegraphics[width=0.48\textwidth]{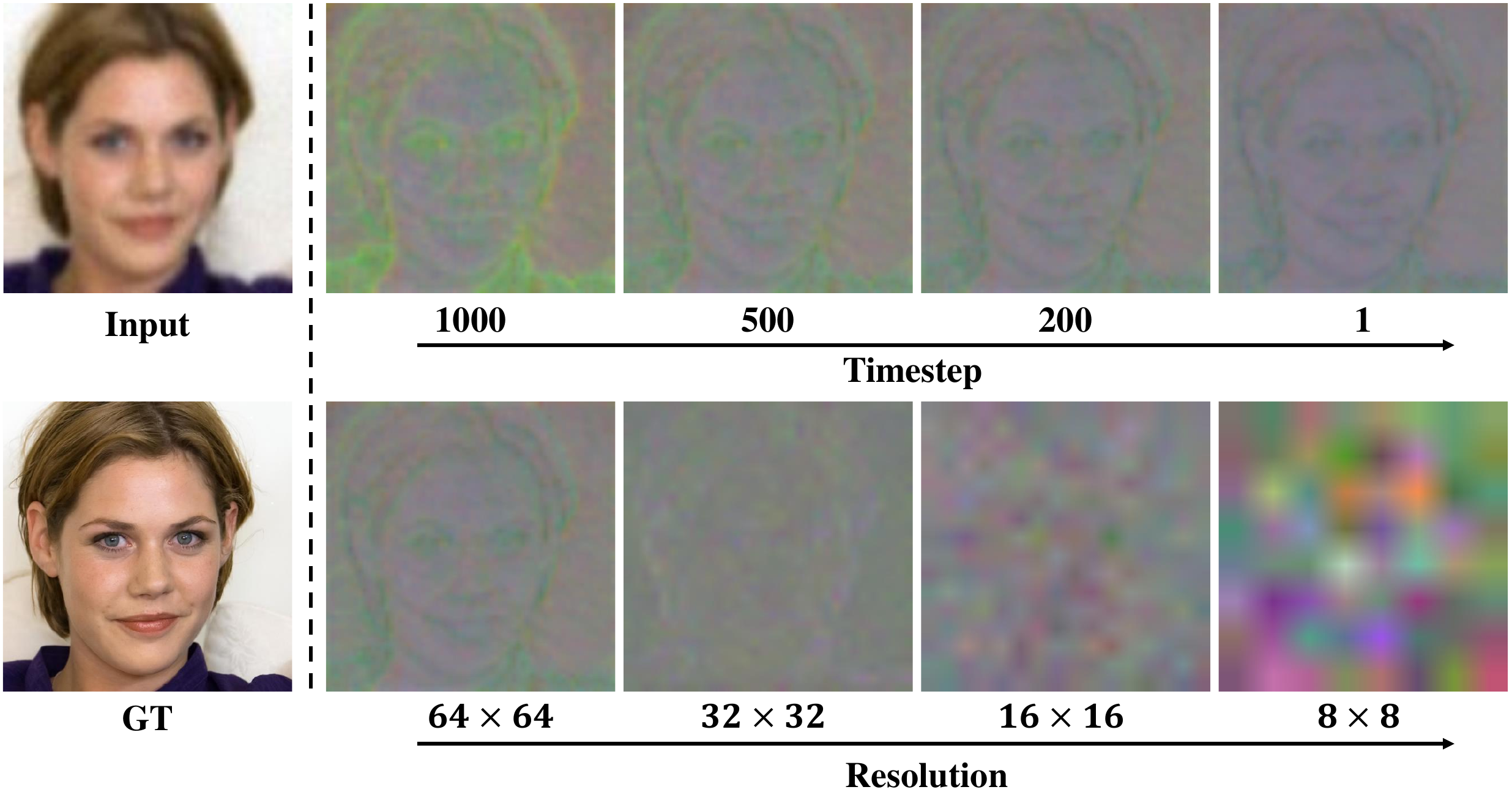}
\caption{Visualization of feature maps learned by our multi-scale feature extraction module (MFEM) in different timesteps and resolutions. The first row demonstrates the capability of our MFEM to extract accurate features at any timesteps. The second row shows the multi-scale features extracted by our MFEM at various resolutions.}
\label{featuremaps}
\end{figure}

Fig. \ref{featuremaps} shows the visualization of \(F^{n}\) learned by our MFEM in different timesteps and resolutions. Thanks to the delicately designed transformer blocks, our MFEM can extract accurate features at any timesteps. Besides, we find that the feature maps around 200 timesteps are clearer, and such phenomena have similarly occurred in previous works \cite{stalesr}. The second row shows the multi-scale features that match the different resolutions of Stable Diffusion blocks. The proposed MFEM constitutes one of the essential prerequisites enabling our method to achieve state-of-the-art performance.

\subsubsection{Trainable Time-aware Prompt Module}
Stable Diffusion 2.1-base is a large text-to-image diffusion model, which encodes the text information into vectors using a fixed, pretrained OpenCLIP-ViT/H text encoder with $354$M parameters. Previous works \cite{stalesr, diffbir} that utilize Stable Diffusion as the prior in Low-level Vision tasks usually set the prompt as \textbf{null} to generate the fixed latent vectors. However, there are two disadvantages by doing so: 1) The fixed CLIP \cite{CLIP} text encoder is so large that it requires a lot of computational resources during both the training and inference stages. 2) The operation is not beneficial to the tasks at all. 

To address these two problems, we propose the trainable time-aware prompt module (TTPM) to generate latent prompts that can guide the restoration process in different time steps. Specifically, our prompt component is a trainable parameter P \(\in \mathbb{R} ^{77\times 1024} \), which matches the size of the output of the CLIP text encoder. In order to generate more effective prompts in different time steps, we further incorporate the time embedding into the prompt via a cross-attention layer. The overall process of TTPM is defined as:
\begin{equation}
Prompt = MLP(CrossAttention(P,emb) + P),
\label{eq11}
\end{equation}
where \(emb\) represents the time embedding and \(Prompt\) is the output of TTPM that can provide semantic guidance for the restoration process at each time step.

\begin{table}[t]
    \centering
    \caption{The computational complexity between our BFRffusion and other diffusion-based methods}
    \begin{tabular}{c|ccc}
    \hline
        Method & Parameters & MACs & Time(s/image) \\ \hline
        StableSR \cite{stalesr} & \textbf{1.20B} & 1498.52G & 1.84 \\ 
        DiffBIR \cite{diffbir} & 1.51B & 1632.51G & 2.97 \\ 
        \textbf{BFRffusion(ours)} & 1.23B & \textbf{986.38G} & \textbf{1.45} \\ \hline
    \end{tabular}
    \label{complexity}
\end{table}

The computational complexity between our BFRffusion and other diffusion-based methods is shown in Table \ref{complexity}. We utilize the profile function from the thop package to compute the parameter count and Multiply-Accumulate Operations (MACs). The inference time is conducted on 1 NVIDIA A100 GPU. Thanks to our TTPM, our method significantly reduces computational complexity and holds an advantage in inference time.

\subsubsection{Pretrained Denoising U-Net Module}
We finetune the pretrained denoising U-Net of Stable Diffusion to build our restoration network BFRffusion. In order to improve the sampling efficiency and reduce the computational requirements, Stable Diffusion applies a pretrained variational autoencoder (VAE) model with KL loss to transform the diffusion and denoising process from pixel space to latent space. During the diffusion process, the latent image \(z\) which is encoded from the pixel image \(x\) using the VAE, is nearly transformed to a standard Gaussian noise \(z_{t}\) by adding random noise gradually. With the help of the reparameterization trick, we can directly calculate \(z_{t}\) based on the initial latent image \(z\) and the time step \(t\). The process is calculated as follows:
\begin{equation}
\bar{\alpha}_{t} = {\textstyle \prod_{i=1}^{t}} \alpha_{i} = {\textstyle \prod_{i=1}^{t}}(1 - \beta_{i}),
\label{eq12}
\end{equation}
\begin{equation}
z_{t}=\sqrt{\bar{\alpha}_{t}} z+\sqrt{1-\bar{\alpha}_{t}} \epsilon,
\label{eq13}
\end{equation}
where \(\beta_{i}\) represents the variance of the added Gaussian noise at the \(i^{th}\) time step, \(\epsilon\) is a random Gaussian noise and \(\epsilon \sim \mathcal{N} (0,I)\). The reverse process generates clear samples from random noise \(z_{T} \sim \mathcal{N} (0, I)\) by denoising gradually. Our BFRffusion adopts the pretrained U-Net of Stable Diffusion as the denoiser. We further add the output features \(F^{n}\) from our MFEM to the denoiser to guide the denoising process. The \(Prompt\) from our TTPM is mapped to the denoiser via a cross-attention layer to provide semantic guidance. The optimization of our BFRffusion is defined as follows,
\begin{equation}
\mathcal{L} = \mathbb{E} _{z_{t},t,F^{n},Prompt} [ || \epsilon -\epsilon _{\theta}(  z_{t},t,F^{n},Prompt)||_{2}^{2} ],
\label{eq14}
\end{equation}
where \(\epsilon _{\theta}\) is the U-Net denoiser, and \(t\) is the time steps of adding noise.

\subsection{Privacy-preserving Face Dataset for Blind Face Restoration}

In this section, we provide an overview of our PFHQ dataset and introduce how it is built in detail. As mentioned earlier, we aim to build a privacy-preserving and balanced face dataset, which provides paired high-quality and low-quality face images. We expect this dataset could drive the development of the blind face restoration task in the future.

The face dataset construction consists of the following steps: selection of an appropriate face image generation model, generation of face images, selection and classification of face images, and synthesis of paired high-quality and low-quality images.

\subsubsection{Stage I: Choose an Appropriate Face Image Generation Model}
\begin{figure}[t] 
\centering
\includegraphics[width=0.48\textwidth]{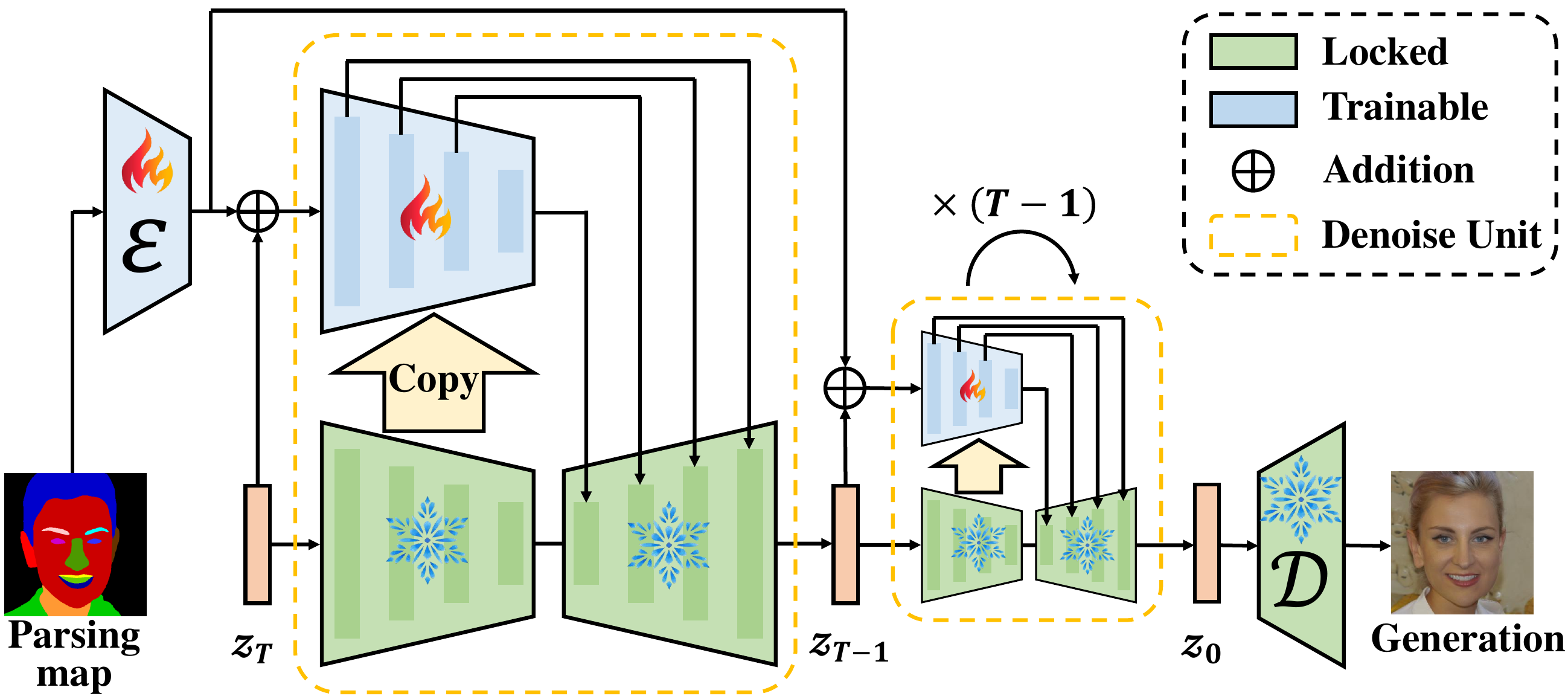}
\caption{The pipeline of our face image generation process. We choose aligned face parsing maps as the input of the pipeline.}
\label{ControlNet}
\end{figure}

The mainstream generative models today include Generative Adversarial Networks (GANs) \cite{gan} and Diffusion Models. In 2018, Karras \etal \cite{progressivegan} propose a novel GAN training methodology that allows for progressive growth of both the generator and discriminator. This methodology makes it possible to generate high-resolution face images. Subsequently, StyleGAN \cite{ffhq} and StyleGAN2 \cite{stylegan2} further improve the quality of face image generation through changes in both model architecture and training methods. Although GAN-based generation models could produce realistic images, they often exhibit a deficiency in diversity. Recently, research \cite{DDPM, beatgans, improveddiffusion} shows that diffusion models can achieve superior image quality and diversity compared to the GAN-based models. In order to reduce computational requirements, the latent diffusion model \cite{latentdiffusion} applies diffusion and denoising process in the latent space using a pretrained autoencoder, and Stable Diffusion is a large-scale implementation of it. While Stable Diffusion is a powerful text-to-image generation model, only text is insufficient for generating a large number of high-quality aligned face images. ControlNet \cite{ControlNet} can add spatial conditioning control to the pretrained Stable Diffusion and generate various face images. Therefore, we choose ControlNet as the face image generation architecture. Additionally, we choose face parsing maps as the additional spatial conditioning control for ControlNet. Face parsing maps can provide a pixel-wise description of the various parts of a person's face. The pipeline of our image generation process is shown in Fig. \ref{ControlNet}.

\subsubsection{Stage II: Generate the Face Images}
\begin{figure}[t] 
\centering
\includegraphics[width=0.48\textwidth]{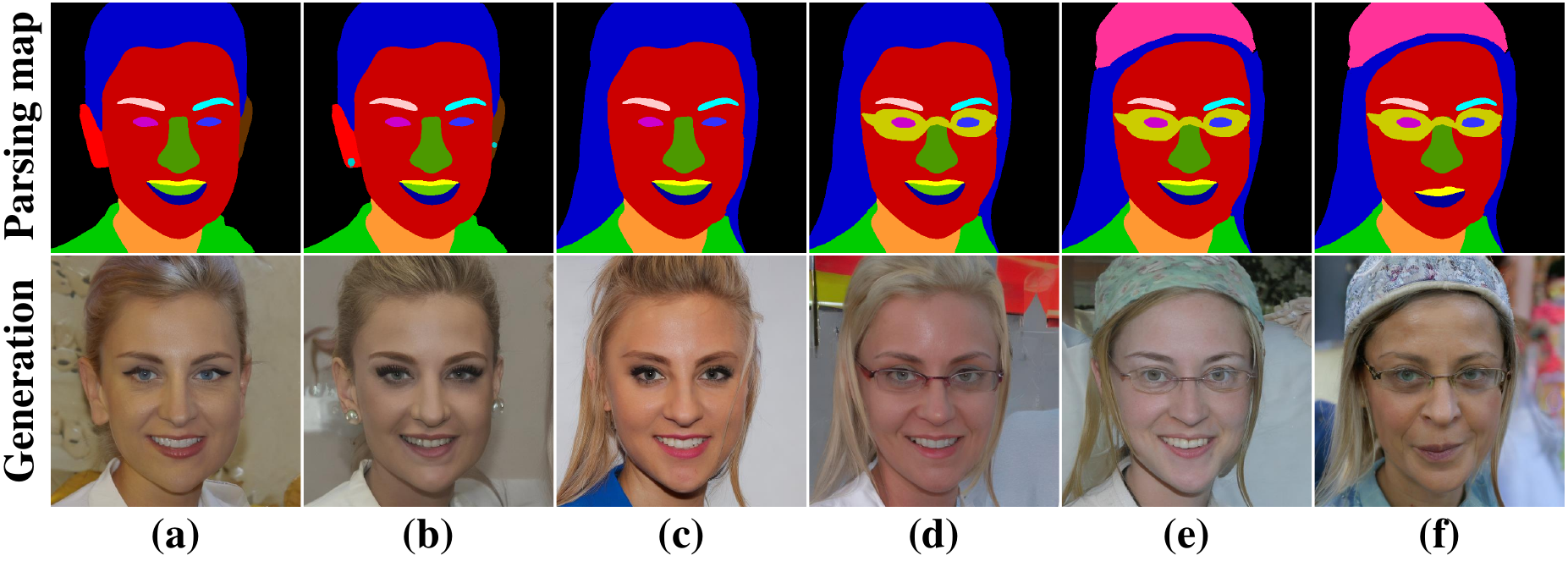}
\caption{Visual results of modification to the face parsing maps. The first row shows examples of the face parsing maps and the second row shows corresponding image generation results. The modifications are as follows: (a) base, (b) adding earrings, (c) changing the hairstyle, (d) adding glasses, (e) adding a hat, (f) changing mouth style. \textbf{Zoom in for best view.}}
\label{mask}
\end{figure}

We train ControlNet \cite{ControlNet} on a combined dataset comprising 70K face images from the FFHQ dataset \cite{ffhq} and 27K face images from the training set of the CelebA-HQ dataset \cite{progressivegan}. The training has no overlap with the testing datasets in our evaluation stage. Subsequently, all the high-quality images are resized to \(512\times 512\). To acquire the face parsing maps, a pretrained face parsing network \cite{PSFRGAN} is employed to synthesize these maps from high-quality images. The ControlNet is trained using the paired high-quality images and face parsing maps through 150k iterations with a learning rate of \(1\times10^{-4}\). In the inference stage, we set the sample step to $50$ and utilize the face parsing maps generated by a pretrained unconditional generation model \cite{latentdiffusion} as the input maps to the ControlNet. Fig. \ref{mask} visually illustrates how slight modifications to the face parsing maps enable the generation of diverse and realistic face images. We generate a large number of face images for further selection and classification in the subsequent stage.

\subsubsection{Stage III: Select and Classify the Face Images}
Utilizing OpenCV, we measure the sharpness of synthetic face images via Laplacian variance and choose images with a sharpness surpassing 150 for subsequent classification. Fairface \cite{fairface} is an artificially annotated dataset with balanced race, gender, and age. The prediction model trained on the Fairface dataset could achieve better classification accuracy. Thus, we employ this prediction model to classify the race, gender, and age of the selected images. We obtain four race groups: Asian, Black, White, and Other (including Indian, Latino, \etc), two gender groups: male and female, and six age groups: 0-9, 10-19, 20-29, 30-39, 40-49 and 50+ years old. We randomly choose 1,250 and 10 face images to build the training and testing dataset from each distinct race, gender, and age group.

\subsubsection{Stage IV: Synthesize Pairs of Images}
Through the described procedures, we have obtained 60K and 480 high-quality face images that are both privacy-preserving and balanced. In order to fulfill the training and testing requirements of current blind face restoration methods, we synthesize low-quality images from the corresponding high-quality images. We employ the degeneration strategies employed in most existing blind face restoration methods \cite{SCGAN, GFRNet, DFDNet, PSFRGAN, GFPGAN, VQFR, Restoreformer, codeformer, diffbir}. To be specific, we employ the following formula to synthesize the low-quality images from the high-quality images:
\begin{equation}
y=\left[\left(x \circledast k_{\sigma}\right) \downarrow_{r}+ n_{\delta}\right]_{\mathrm{JPEG}_{q}},
\label{eq15}
\end{equation}
where  \(y\) is the low-quality images, \(x\) is the high-quality images, \(\circledast\) represents the convolution operation, \(k_{\sigma}\) is Gaussian blur kernel, \(\downarrow_{r}\) represents downsampling, \(n_{\delta}\) is white Gaussian noise and \(\mathrm{JPEG}_{q}\) represents JPEG compression. The factors \(\sigma\), \(r\), \(n\), and \(q\) are parameters of the above operations and are randomly sampled from \(\left \{0.2:10\right \} \), \(\left \{1:8\right \} \), \(\left \{0:15\right \} \), \(\left \{60:100\right \} \), which are consistent with previous works \cite{GFPGAN, VQFR}. 

In summary, by conducting the above four steps, we obtain a synthetic and balanced training dataset comprising 60K pairs of face images that can be used for training blind face restoration methods. We use the other 480 pairs of face images to build the PFHQ-Test to evaluate the performance of restoration networks on balanced data.

\begin{figure*}[t]
\centering
\includegraphics[width=1\textwidth]{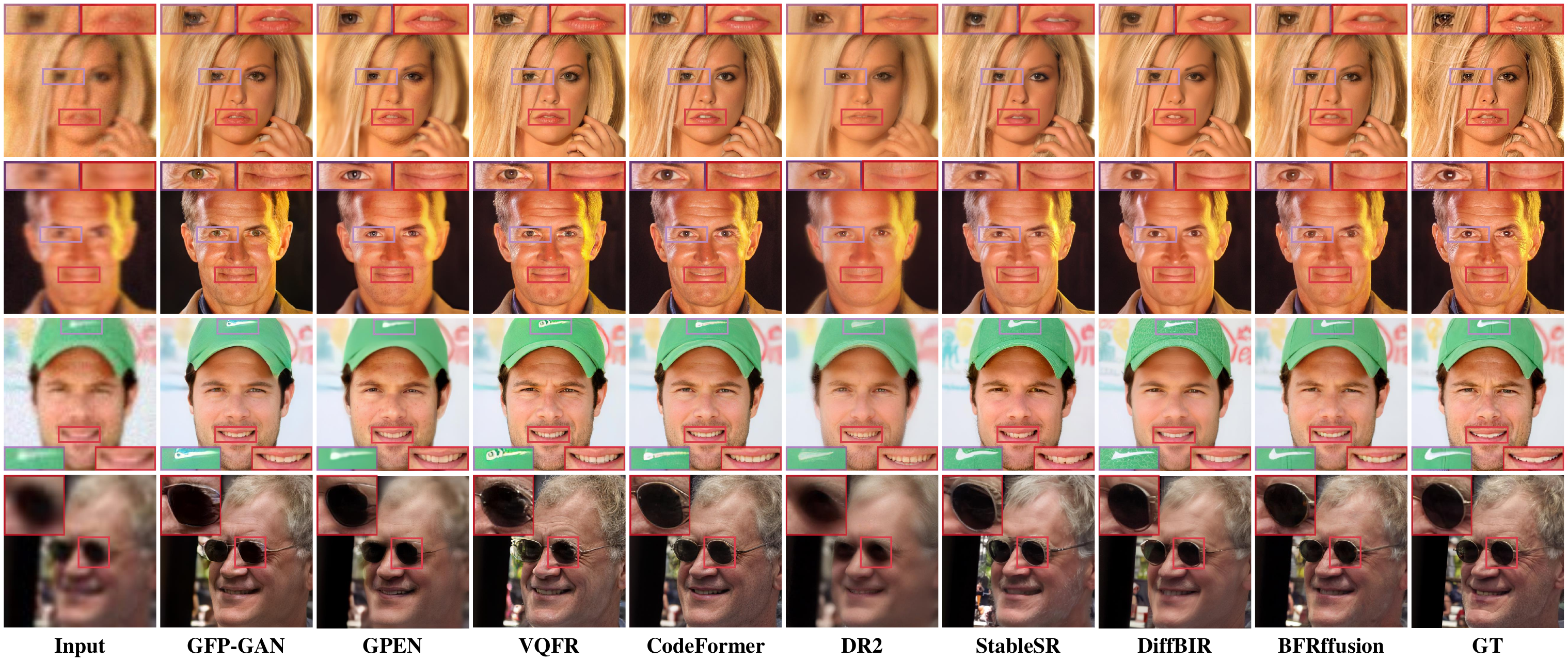}
\caption{Visual comparison results of different methods on the \textbf{CelebA-Test}. Our BFRffusion produces more faithful details. \textbf{Zoom in for best view.}}
\label{celeba_compare}
\end{figure*}

\section{Experiment} \label{Experiment}

\subsection{Implementation, Datasets and Metrics Details}
\subsubsection{Implementation} Stable Diffusion 2.1-base is adopted as our foundational denoising networks and generative facial prior. We employ a new training strategy compared to the strategy used in StableSR \cite{stalesr} and DiffBIR \cite{diffbir}, which maintain the U-Net frozen throughout all training phases. We first finetune the frozen diffusion model for 100K iterations. Subsequently, we unfreeze the decoder weights of the U-Net in Stable Diffusion and train the whole restoration model for 150K iterations. The adamW optimizer (\(\beta_1 = 0.9, \beta_2=0.999\), weight decay = 0.01) is employed with the cosine annealing strategy, where the learning rate gradually decreases from the initial learning rate \(1\times10^{-4}\) to zero for the last 50K iterations. We set the batch size to 64 on 4 NVIDIA A100 GPUs in the training stage and apply the DDIM \cite{DDIM} sampler with 50 steps in the inference stage.
\subsubsection{Datasets} We train our BFRffusion on the dataset with paired high-quality and low-quality images. Specifically, the high-quality images are 70K images of the FFHQ dataset \cite{ffhq}, which are resized to \(512\times 512\). Following the previous works \cite{SCGAN, GFRNet, DFDNet, PSFRGAN, GFPGAN, VQFR, Restoreformer, codeformer, diffbir}, we synthesize the low-quality images from the high-quality images by the degrading formula Eq. \eqref{eq15}. And the factors in the formula are identical to those in previous works \cite{GFPGAN, VQFR}. In order to improve the efficiency of training, the low-quality images are also resized to \(512\times 512\). 

Similar to \cite{GFPGAN, VQFR, Restoreformer, codeformer, diffbir}, we employ a synthetic paired dataset CelebA-Test and three real-world datasets: LFW-Test, CelebAdult-Test, and WIDER-Test to evaluate our BFRffusion. CelebA-Test consists of 3K paired images that are synthesized from CelebA-HQ \cite{progressivegan} images. LFW-Test comprises 1,711 low-quality images which are the first images for each identity in the validation partition of LFW. CelebAdult-Test which consists of 180 adult faces is collected from the Internet. WIDER-Test consists of 970 severely degraded face images from the WIDER Face dataset \cite{yang2016wider}. All these testing datasets are public and do not overlap with the training dataset.

\subsubsection{Metrics} 
For the paired testing dataset CelebA-Test, we employ both pixel-wise metrics (PSNR and SSIM) and the perceptual metrics (LPIPS \cite{lpips} and FID \cite{FID}) to evaluate our BFRffusion. Specifically, PSNR is defined via the mean squared error (MSE) of pixels and SSIM focuses on the structural information (e.g., brightness, contrast, \etc.) of images. LPIPS \cite{lpips} extracts features from the images and then calculates the perceptual difference between these features using a pretrained VGG network. FID \cite{FID} calculates the similarity of feature vectors extracted from a pretrained inception model between the training dataset and the restored images. Additionally, we use 'Deg.' to denote the identity metric, measuring the identity distance through angles within the ArcFace \cite{arcface} feature embedding. For assessing the fidelity with accurate facial positions and expressions, we employ the landmark distance (LMD) as the fidelity metric. For the real-world testing datasets, we adopt the widely-used non-reference metric FID \cite{FID}.

\subsection{Comparison with SOTA Face Restoration Methods} \label{IVB}
\begin{table}[t]
    \caption{Quantitative comparison on \textbf{CelebA-Test} for blind face restoration. {\color{red}\textbf{Red}} and {\color{blue}\underline{Blue}} indicate the best and the second-best performance.}
    \centering
    \tabcolsep=0.1cm
    \begin{tabular}{c|cc|cc|cc}
    \hline
        Method & LPIPS$\downarrow$ & FID$\downarrow$ & Deg.$\downarrow$ & LMD$\downarrow$ & PSNR$\uparrow$ & SSIM$\uparrow$ \\ \hline
        Input & 0.4866 & 143.98 & 47.94 & 3.76 & 25.35 & 0.6848 \\ 
        HiFaceGAN \cite{hifacegan} & 0.4770 & 66.09 & 42.18 & 3.16 & 24.92 & 0.6195 \\ 
        DFDNet \cite{DFDNet} & 0.4341 & 59.08 & 40.31 & 3.31 & 23.68 & 0.6622 \\ 
        PSFRGAN \cite{PSFRGAN} & 0.4240 & 47.59 & 39.69 & 3.41 & 24.71 & 0.6557 \\ \hline
        GFP-GAN \cite{GFPGAN} & 0.3646 & 42.62 & 34.60 & 2.41 & 25.08 & 0.6777 \\ 
        GPEN \cite{GPEN} & 0.4009 & \color{red}\textbf{36.46} & 35.54 & 2.64 & 25.59 & {\color{blue}\underline{0.6894}} \\ \hline
        VQFR \cite{VQFR} & {\color{blue}\underline{0.3515}} & 41.28 & 35.75 & 2.43 & 24.14 & 0.6360 \\ 
        CodeFormer \cite{codeformer} & {\color{red}\textbf{0.3432}} & 52.43 & {\color{blue}\underline{32.51}} & {\color{blue}\underline{2.12}} & 25.15 & 0.6699 \\ \hline
        DR2 \cite{wang2023dr2} & 0.3979 & 58.94 & 48.30 & 3.34 & 24.44 & 0.6784 \\
        StableSR \cite{stalesr} & 0.3637 & {\color{blue}\underline{39.73}} & 36.29 & 2.36 & 24.84 & 0.6772 \\
        DiffBIR \cite{diffbir} & 0.3786 & 47.90 &  35.02 & 2.24 & {\color{blue}\underline{25.60}} & 0.6809 \\ 
        \textbf{BFRffusion (ours)} & 0.3621 & 40.74 & {\color{red}\textbf{30.91}} & {\color{red}\textbf{1.99}} & {\color{red}\textbf{26.20}} & {\color{red}\textbf{0.6926}} \\ \hline
    \end{tabular}
    \label{CelebA-Test}
\end{table}

\begin{figure*}[t]
\centering
\includegraphics[width=1\textwidth]{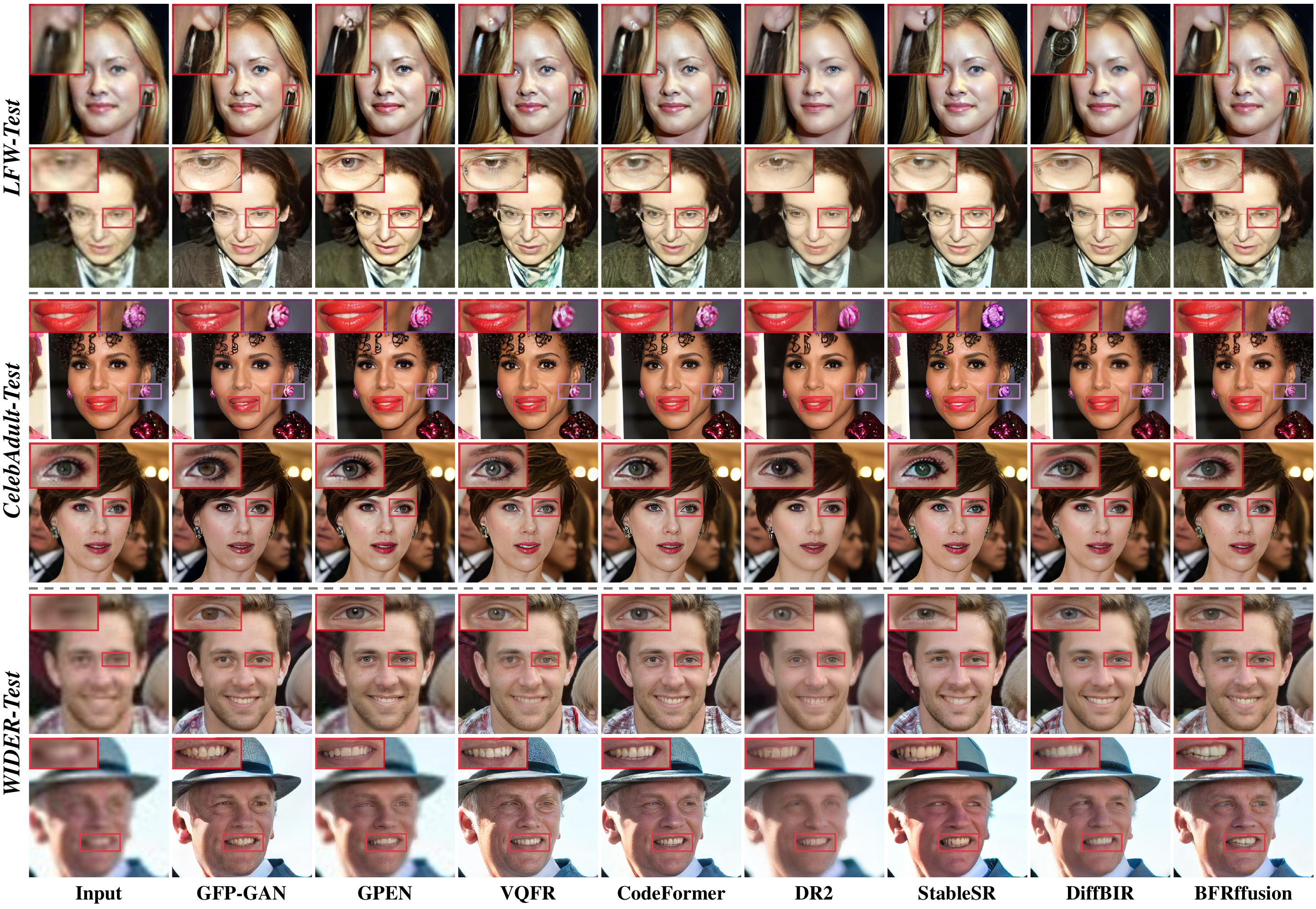}
\caption{Visual comparison results of different methods on three \textbf{real-world} datasets. Our BFRffusion produces more faithful details. \textbf{Zoom in for best view.}}
\label{realworld_compare}
\end{figure*}

We compare the performance of our BFRffusion with several state-of-the-art face restoration methods: HiFaceGAN \cite{hifacegan}, DFDNet \cite{DFDNet}, PSFRGAN \cite{PSFRGAN}, GFP-GAN \cite{GFPGAN}, GPEN \cite{GPEN}, VQFR \cite{VQFR}, CodeFormer \cite{codeformer}, DR2 \cite{wang2023dr2}, StableSR \cite{stalesr} and DiffBIR \cite{diffbir}. Specifically, HiFaceGAN \cite{hifacegan} formulates the face restoration task as a generative problem guided by semantics, and the problem is tackled by a multi-stage framework consisting of collaborative suppression and replenishment. DFDNet \cite{DFDNet} generates a deep dictionary of key facial components extracted from high-quality images as the reference prior and selects similar features to degraded inputs to guide the restoration process. PSFRGAN \cite{PSFRGAN} is a multi-scale progressive restoration network, that leverages both the geometric prior (parsing maps) and pixel space information (input degraded images). GFP-GAN \cite{GFPGAN} employs the pretrained StyleGAN2 \cite{stylegan2}, which encapsulates rich and diverse generative facial priors, and spatial feature transform layers to restore realistic and faithful faces. GPEN \cite{GPEN} is a two-stage framework, which first embeds learned GAN blocks into a U-shaped DNN as a prior decoder, and then fine-tunes the GAN prior with the synthesized low-quality face images. VQFR \cite{VQFR} is a VQ-based face restoration method equipped with the VQ codebook as a facial detail dictionary and the parallel decoder design. 
CodeFormer \cite{codeformer} is a Transformer-based prediction network, that models both the global composition and the context of low-quality faces for code prediction. It has shown superior robustness to degradation with the codebook prior. DR2 \cite{wang2023dr2} captures semantic information through iterative denoising steps and is robust against common degradation. StableSR \cite{stalesr} is a blind super-resolution method leveraging prior knowledge encapsulated in pretrained Stable Diffusion. DiffBIR \cite{diffbir} is a two-stage framework with the pretrained Stable Diffusion.
In summary, our selection of the above methods is guided by the three following criteria. First, these methods achieve state-of-the-art performance in terms of widely used metrics such as PSNR, SSIM, \etc. Second, the testing source codes of these methods are publicly accessible. Third, these methods are specifically designed for blind face restoration. 

\begin{table}[t]
    \caption{Quantitative comparison on three \textbf{real-world} datasets for blind face restoration. {\color{red}\textbf{Red}} and {\color{blue}\underline{blue}} indicate the best and the second-best performance.}
    \centering
    \tabcolsep=0.1cm
    \begin{tabular}{c|ccc}
    \hline
        Dataset & \textbf{LFW-Test} & \textbf{CelebAdult-Test}  & \textbf{WIDER-Test}\\
        Method & FID$\downarrow$ & FID$\downarrow$ & FID$\downarrow$ \\ \hline
        Input &  137.56 & 118.33  & 202.06 \\ 
        HiFaceGAN \cite{hifacegan} &  64.50 & {\color{blue}\underline{104.02}}  & 125.56 \\ 
        DFDNet \cite{DFDNet} & 62.57 & 105.71   & 57.84 \\ 
        PSFRGAN \cite{PSFRGAN} &  51.89 & 104.08  & 51.16 \\ \hline
        GFP-GAN \cite{GFPGAN} &  49.96 & 104.75   & 40.59 \\ 
        GPEN \cite{GPEN} & 57.00 & 107.53   & 46.99 \\ \hline
        VQFR \cite{VQFR} &  50.64 & 104.99  & 37.87 \\ 
        CodeFormer \cite{codeformer} &  52.36 & 110.09  & 39.06 \\ \hline
        DR2 \cite{wang2023dr2} & 49.85 & 107.70 & 54.34  \\
        StableSR \cite{stalesr} & {\color{blue}\underline{41.15}} & 105.09 & {\color{blue}\underline{33.05}}  \\
        DiffBIR \cite{diffbir} &  {\color{red}\textbf{39.58}} & 105.02  & {\color{red}\textbf{32.35}} \\ 
        \textbf{BFRffusion (ours)} &  49.92 & {\color{red}\textbf{103.64}}  & 56.97 \\ \hline
    \end{tabular}
    \label{Realworld-Test}
    \vspace{-4mm}
\end{table}

\subsubsection{Comparisons on Synthetic Dataset}
The quantitative comparisons of the methods mentioned above are reported in Table \ref{CelebA-Test}. The results show that our BFRffusion achieves state-of-the-art performance on the CelebA-Test. Specifically, our BFRffusion achieves the best performance regarding pixel-wise metrics PSNR and SSIM. Furthermore, BFRffusion obtains the lowest Deg. and LDM scores, indicating its ability to accurately recover identity and facial details. In addition, BFRffusion achieves comparable LPIPS and FID scores, suggesting that the quality of restored faces is close to the ground truth.
The qualitative results are presented in Fig. \ref{celeba_compare}. Our BFRffusion leverages the proposed multi-scale feature extraction module to extract image features during the restoration process, allowing it to capture clear details of the entire image. In contrast, the methods, that require face detection (DFDNet \cite{DFDNet}) or parsing maps (PSFRGAN \cite{PSFRGAN}), are difficult to restore faithful details in the whole image. Moreover, thanks to the remarkable generative ability of the pretrained Stable Diffusion, our BFRffusion successfully recovers realistic details in the eyes, mouth, decorations, \etc. On the contrary, GAN-based methods \cite{GFPGAN, GPEN, VQFR, codeformer} are difficult to restore complex components, due to their limited generative ability. In conclusion, our BFRffusion performs better in the realness and fidelity of the facial details.

\begin{figure*}[t]
\centering
\includegraphics[width=1\textwidth]{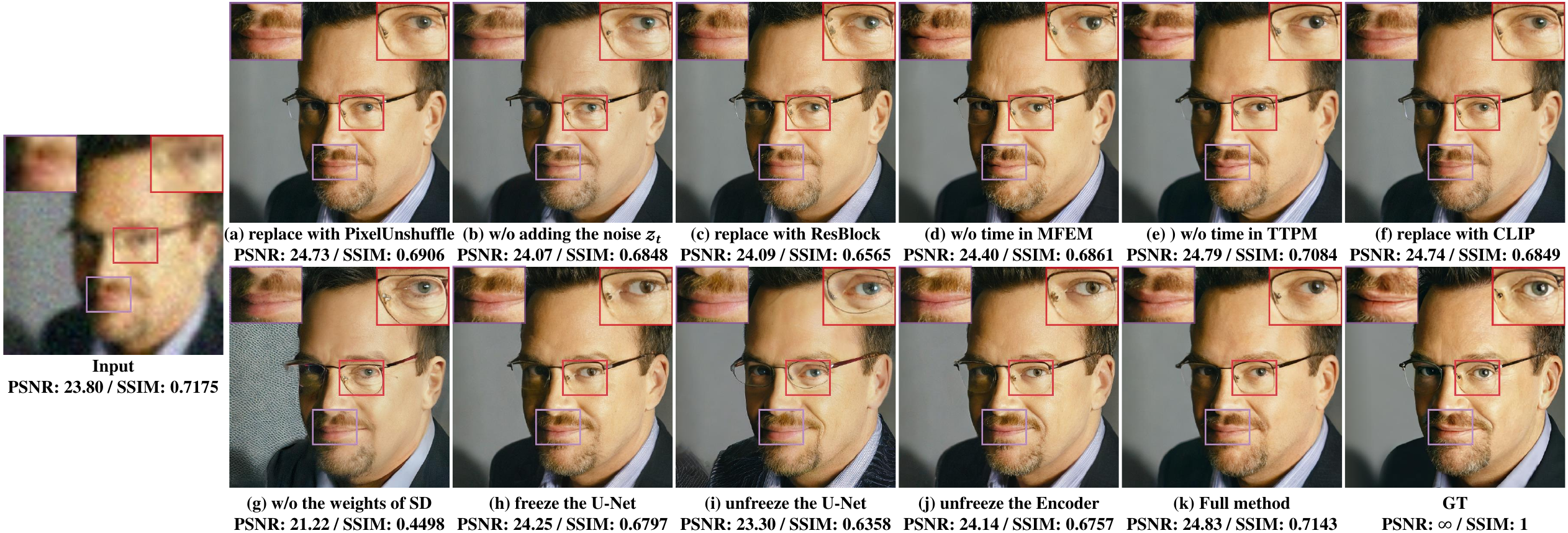}
\caption{Visual comparison results of ablation studies on BFRffusion. The internal modules of our BFRffusion and the strategies employed during training all play important roles in the effectiveness of the restoration process.}
\label{Ablationfig}
\end{figure*}

\subsubsection{Comparisons on Real-world Datasets}
Furthermore, we compare the performance of the methods mentioned above in real-world scenarios based on three real-world datasets to evaluate the generalization ability. The quantitative comparisons are shown in Table \ref{Realworld-Test}. Our BFRffusion achieves the best performance on the CelebAdult-Test datasets, while also demonstrating comparable performance on the LFW-Test. However, our BFRffusion achieves poor performance on the WIDER-Test with severe degradation. We use up to 8x downsampling to simulate low-resolution situations and our model struggles to restore such highly degraded images in WIDER-Test. The qualitative results are presented in Fig. \ref{realworld_compare}. Our BFRffusion equipped with powerful generative prior is able to generate more realistic and detailed features such as teeth, eyes, makeup, and decorations. Although DiffBIR \cite{diffbir} obtains the lowest FID scores on the LFW-Test and WIDER-Test, it notably lacks fine facial texture details and tends to appear smoother as shown in Fig. \ref{realworld_compare}.


\begin{table}[!ht]
    \caption{Ablation Studies of our BFRffusion}
    \centering
    \tabcolsep=0.20cm
    \begin{tabular}{c|p{0.15cm} c|c c}
    \hline
        Module & \multicolumn{2}{c|}{Configuration}  & PSNR$\uparrow$ & SSIM$\uparrow$ \\ \hline
        \multirow{2}*{SDRM} & \textbf{\textcolor{blue}{(a)}} & replace with PixelUnshuffle \cite{pixelunshuffle} & 26.06 & 0.6844 \\ 
		~ & \textbf{\textcolor{blue}{(b)}} & w/o adding the noise \(z_{t} \) & 24.66 & 0.6581 \\\hline
        \multirow{2}*{MFEM} & \textbf{\textcolor{blue}{(c)}} & replace with ResBlock \cite{residual} & 25.57 & 0.6608 \\ 
		~ & \textbf{\textcolor{blue}{(d)}} & w/o time embedding & 25.48 & 0.6569 \\\hline
        \multirow{2}*{TTPM} & \textbf{\textcolor{blue}{(e)}} & w/o time embedding & 26.03 & 0.6832 \\
            ~ & \textbf{\textcolor{blue}{(f)}} & replace with CLIP \cite{CLIP} & 25.92 & 0.6768 \\\hline
        \multirow{4}*{PDUM} & \textbf{\textcolor{blue}{(g)}} & w/o the weights of SD & 24.31 & 0.6063 \\ 
            ~ & \textbf{\textcolor{blue}{(h)}} & freeze the U-Net \cite{unet}  & 25.65 & 0.6615 \\ 
		~ & \textbf{\textcolor{blue}{(i)}} & unfreeze the U-Net & 24.75 & 0.6272 \\ 
            ~ & \textbf{\textcolor{blue}{(j)}} & unfreeze the Encoder & 25.50 & 0.6557 \\ \hline
        Overall & \textbf{\textcolor{blue}{(k)}} & Full method & \textbf{26.20} & \textbf{0.6926} \\ \hline
    \end{tabular}
    \label{Ablation}
\end{table}

\subsection{Ablation Studies}
In this section, we analyze and discuss the effectiveness of the internal modules in our BFRffusion and the strategies employed during our training process. All models are trained on FFHQ \cite{ffhq} and tested on CelebA-Test. The quantitative results are shown in Table \ref{Ablation} and the qualitative results are presented in Fig. \ref{Ablationfig}.

\subsubsection{Shallow Degradation Removal Module}
There are two main functions of the proposed shallow degradation removal module (SDRM): (1) remove shallow degradation and encode pixel images into latent images using an encoder network. (2) add untreated noise of different diffusion steps to latent images. Firstly, we replace the encoder network with PixelUnshuffle operation, leading to a slight decrease in performance. It implies that our encoder network can remove shallow degradation, while the PixelUnshuffle operation just changes the pixel arrangement of input images. Secondly, we remove the untreated noise added to the latent images, which results in a significant performance drop. It indicates that the noise is important for the restoration process because of the characteristics of diffusion models.

\subsubsection{multi-scale Feature Extraction Module}
Our multi-scale feature extraction module (MFEM) is constructed using transformer blocks with time conditions. To test the importance of the transformer blocks, we replaced them with time-embedded ResBlocks \cite{resblock, residual},  which caused a significant drop in performance. It suggests that the proposed transformer blocks are important for the extraction of global information and local information. Then we remove the time conditions of the transformer blocks. This also causes a performance drop, demonstrating that time information is necessary for the feature extraction and denoising process.

\subsubsection{Trainable Time-aware Prompt Module}
The trainable time-aware prompt module (TTPM) consists of a trainable prompt and time embedding. Firstly, we remove the time conditions, and there is only a trainable prompt left. Then we replace TTPM with the frozen CLIP text encoder and set the input prompt to blank, which is equivalent to a fixed prompt and is similar to previous works \cite{stalesr, diffbir}. Table \ref{Ablation}{\textcolor{blue}{(e)}} shows that our TTPM with the time conditions yields 0.17 dB PSNR gain over only the trainable prompt. What's more, Table \ref{Ablation}{\textcolor{blue}{(f)}} shows that the trainable prompt yields 0.11 dB PSNR gain over the fixed prompt.

\subsubsection{Pretrained Denoising U-Net Module}
We adopt the pretrained denoising U-Net module (PDUM) of Stable Diffusion as the foundation denoising networks and generative facial prior. To demonstrate the effectiveness of the pretrained Stable Diffusion weights, we train our BFRffusion without them for initializing the U-Net denoiser. A significant drop is observed and it can't generate realistic glasses as shown in Fig. \ref{Ablationfig}(g), which shows that the generative ability from the pretrained Stable Diffusion is quite important for the restoration process. We employ a training strategy where we finetune the frozen diffusion model for 100K iterations, then unfreeze the decoder weights of the U-Net in Stable Diffusion and train the whole restoration model for 150K iterations. This training strategy allows our BFRffusion to achieve wonderful realness and fidelity while retaining the generation ability of Stable Diffusion. We freeze the whole U-Net all the time leading to a bad optimization of the restoration model and a slight performance drop. On the other hand, unfreezing U-Net results in the forgetting of prior knowledge of the pretrained Stable Diffusion and a noticeable performance drop. In the U-Net architecture of diffusion models, the encoder is responsible for gradually downsampling the input into high-dimensional feature representation and extracting rich semantic information, while the decoder gradually restores the spatial details of the image through upsampling operations and performs prediction. We use the proposed MFEM to extract features from the input low-quality images, so We choose to unfreeze the decoder rather than the encoder to achieve better restoration results. Table \ref{Ablation}{\textcolor{blue}{(j)}} shows that the unfrozen decoder yields 0.70 dB PSNR gain over the unfrozen encoder.

\begin{figure}[t] 
\centering
\includegraphics[width=0.48\textwidth]{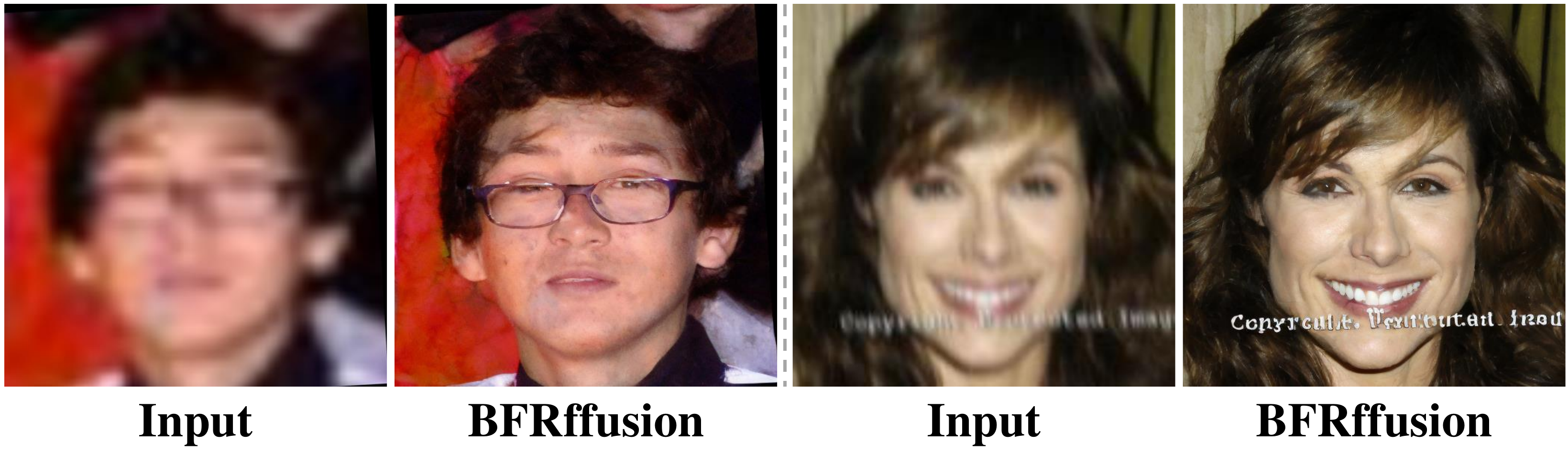}
\caption{Limitations of our BFRffusion. Our BFRffusion struggles to restore clear images when degradation is severe or watermarks are existent.}
\label{Limitations}
\end{figure}

\subsection{Limitations}
Fig. \ref{Limitations} shows some limitations of our BFRffusion. When the degradation of real-world face images is severe, artifacts may appear in the images restored by our BFRffusion. The reason is that our restoration model is trained on the synthetic degraded data, which may not cover all degradation scenarios encountered in the real world. In the future, we plan to utilize higher-quality data to train the restoration model. Additionally, when watermarks appear in face images, our BFRffusion struggles to restore these watermarks while restoring clear faces, significantly affecting the overall visual appearance of the images. It is the existence of these limitations that motivates researchers to continuously explore more effective blind face restoration methods.

\begin{table}[!ht]
    \caption{Quantitative comparison on \textbf{CelebA-Test} and \textbf{PFHQ-Test} for blind face restoration trained on FFHQ and our PFHQ dataset. \textbf{Bold} indicates better performance.}
    \centering
    \begin{tabular}{c|c|c c| c c}
    \hline
        \multirow{2}*{Methods} &Training & \multicolumn{2}{c|}{\textbf{CelebA-Test}}  & \multicolumn{2}{c}{\textbf{PFHQ-Test}} \\ 
        \cline{3-6}
        ~ & Dataset &  PSNR$\uparrow$ & SSIM$\uparrow$ & PSNR$\uparrow$ & SSIM$\uparrow$ \\ \hline
        \multirow{2}*{HiFaceGAN} & FFHQ & 24.92 & 0.6195 & 24.48 & 0.5509\\
            ~ & PFHQ & \textbf{25.04} & \textbf{0.6547} & \textbf{24.60} & \textbf{0.5984} \\ \hline
        \multirow{2}*{PSFRGAN} & FFHQ & 24.71 & \textbf{0.6557} & 24.65 & 0.5784 \\
            ~ & PFHQ & \textbf{25.19} & 0.6519 & \textbf{24.70} & \textbf{0.5923} \\ \hline
        \multirow{2}*{BFRffusion} & FFHQ & \textbf{26.20} & \textbf{0.6926} & 25.24 & 0.6193 \\
            ~ & PFHQ & 26.10 & 0.6876 & \textbf{25.26} &\textbf{ 0.6210} \\ \hline
    \end{tabular}
    \label{Celeba-PFHQ-Test_PFHQ}
\end{table}

\begin{table}[!ht]
    \caption{Quantitative comparison on three real-world datasets for blind face restoration trained on FFHQ and PFHQ dataset. \textbf{Bold} indicates better performance.}
    \centering
    \tabcolsep=0.1cm
    \begin{tabular}{c|c|c c c c c}
    \hline
        \multirow{2}*{Methods} & Training & \textbf{LFW-Test} & \textbf{CelebAdult-Test} & \textbf{WIDER-Test}  \\
            ~ & Dataset & FID$\downarrow$ & FID$\downarrow$ & FID$\downarrow$  \\ \hline
        \multirow{2}*{HiFaceGAN} & FFHQ & 64.50 & \textbf{104.02} & 125.56  \\
            ~ & PFHQ & \textbf{48.00} & 118.88 & \textbf{43.92}  \\ \hline
        \multirow{2}*{PSFRGAN} & FFHQ & 51.89 &\textbf{104.08} & 51.16  \\
            ~ & PFHQ  & \textbf{48.23} & 121.39 & \textbf{43.38} \\ \hline
        \multirow{2}*{BFRffusion} & FFHQ & \textbf{49.92} & \textbf{103.64} & 56.97  \\
            ~ & PFHQ  & 58.19 & 118.10 & \textbf{50.39} \\ \hline
    \end{tabular}
    \label{realworld_PFHQ}
\end{table}

\begin{figure*}[t]
\centering
\includegraphics[width=1\textwidth]{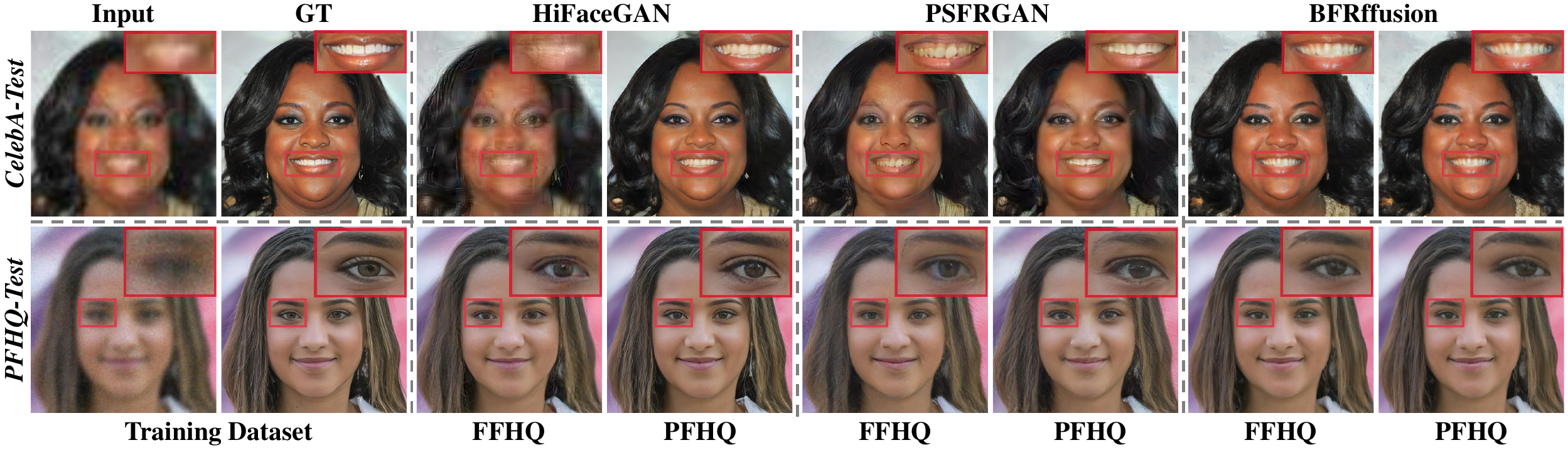}
\caption{Visual comparison results on synthetic testing datasets for methods trained on the FFHQ dataset and PFHQ dataset.}
\label{celeba_PFHQ}
\end{figure*}

\begin{figure*}[t]
\centering
\includegraphics[width=1\textwidth]{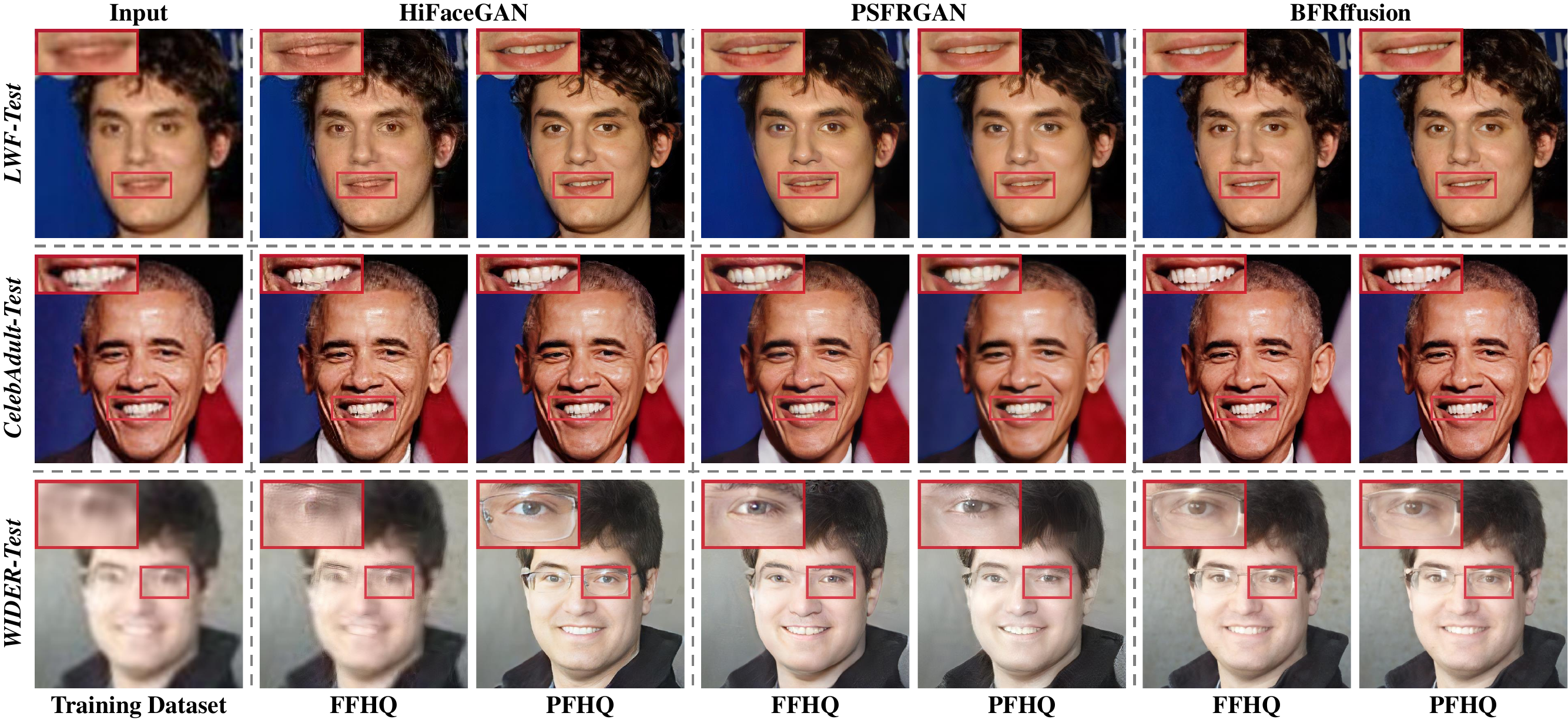}
\caption{Visual comparison results on three real-world testing datasets for methods trained on the FFHQ dataset and PFHQ dataset.}
\label{realworld_PFHQ_figure}
\end{figure*}

\subsection{Utility Evaluation of the Built Privacy-preserving Face Dataset}
To evaluate our privacy-preserving face dataset PFHQ, we train three representative blind face restoration methods: HiFaceGAN \cite{hifacegan} (non-prior-based methods), PSFRGAN \cite{PSFRGAN} (geometric prior-based methods), and BFRffusion (generative prior-based methods) on PFHQ dataset and compare the performance of these methods trained on FFHQ dataset \cite{ffhq}. We test these methods on CelebA-Test, PFHQ-Test, and three real-world testing datasets mentioned in Section \ref{IVB}. The quantitative comparisons on the synthetic dataset CelebA-Test and PFHQ-Test are reported in Table \ref{Celeba-PFHQ-Test_PFHQ}. Bold values indicate better performance. Specifically, On the CelebA-Test, HiFaceGAN \cite{hifacegan} trained on our PFHQ dataset achieves better performance than on the FFHQ dataset. PSFRGAN and BFRffusion trained on our PFHQ dataset achieve comparable performance to these when trained on the FFHQ dataset. On the balanced testing dataset PFHQ-Test, the methods trained on the PFHQ dataset achieve better performance than on the FFHQ dataset. The quantitative comparisons on the public real-world testing datasets are reported in Table \ref{realworld_PFHQ}. Three blind face restoration methods trained on our PFHQ dataset also achieve comparable performance to those trained on the FFHQ dataset. The qualitative results are presented in Fig. \ref{celeba_PFHQ} and \ref{realworld_PFHQ_figure}. Due to the balanced characteristics of our PFHQ dataset, methods trained on it have better applicability to face images of various races, genders, and ages.

The above experiments show that our privacy-preserving face dataset PFHQ achieves comparable performance with the FFHQ dataset for training blind face restoration networks. The generation process of our PFHQ dataset is low-cost and efficient. Furthermore, our dataset can solve privacy issues and reduce the negative effects of racial bias.

\section{Conclusion and Future Work} \label{Conclusion}
In this work, we propose BFRffusion with delicately designed architecture to leverage amazing generative priors encapsulated in pretrained Stable Diffusion for blind face restoration. Our BFRffusion effectively restores realistic and faithful facial details and achieves state-of-the-art performance on both synthetic and real-world public testing datasets. Furthermore, we build a privacy-preserving paired face dataset called PFHQ with balanced race, gender, and age. Extensive experiments show that our PFHQ dataset can serve as an alternative to real face datasets for training blind face restoration methods.

In the future, we plan to address the following challenges in blind face restoration. Firstly, considering the high computational resource consumption of diffusion-based blind face restoration models, it is necessary to devise a low-cost training and inference strategy. Secondly, we plan to explore the potential of synthetic datasets and design more practical synthetic methods for blind face restoration.
\bibliographystyle{IEEEtran}
\bibliography{myreference}

\end{document}